\documentclass[lettersize,journal]{IEEEtran}
\usepackage{amsmath,amsfonts}
\usepackage{algorithmic}
\usepackage{algorithm}
\usepackage{array}
\usepackage[caption=false,font=normalsize,labelfont=sf,textfont=sf]{subfig}
\usepackage{textcomp}
\usepackage{stfloats}
\usepackage{url}
\usepackage{verbatim}
\usepackage{graphicx}
\usepackage{cite}
\usepackage{booktabs}
\usepackage{multirow}
\usepackage{makecell}
\begin{document}

\title{PIER-Flow: Physics-Informed Efficient Rectified Flow for Real-Time Mobile Robot Navigation}

%\author{IEEE Publication Technology,~\IEEEmembership{Staff,~IEEE,}}
	
%\author{IEEE Publication Technology,~\IEEEmembership{Staff,~IEEE,}
%         <-this % stops a space
%\thanks{This paper was produced by the IEEE Publication Technology Group. They are in Piscataway, NJ.}% <-this % stops a space
%\thanks{Manuscript received April 19, 2021; revised August 16, 2021.}}

\author{Shibo~Li, Zhongcheng~Wang, Jiahe~Cao, Jianhua~Yang, and~Ke~Wu%
	\thanks{Manuscript received: Month, Day, Year; Revised: Month, Day, Year; Accepted: Month, Day, Year.}%
	\thanks{Shibo~Li, Zhongcheng~Wang, Jiahe~Cao, and Jianhua~Yang are with the School of Automation, Northwestern Polytechnical University, Xi'an, China (e-mail: lishibo97@mail.nwpu.edu.cn; wangzhongcheng@mail.nwpu.edu.cn; chelde@mail.nwpu.edu.cn; yangjianhua@nwpu.edu.cn). *(Corresponding author: Jianhua Yang.)*}%
	\thanks{Ke~Wu is with the Department of Robotics, Mohamed bin Zayed University of Artificial Intelligence, Abu Dhabi SE45 05, United Arab Emirates (e-mail: ke.wu@mbzuai.ac.ae).}%
	\thanks{Digital Object Identifier (DOI): see top of this page.}%
}

% The paper headers
%\markboth{Journal of \LaTeX\ Class Files,~Vol.~14, No.~8, August~2021}%
%{Shell \MakeLowercase{\textit{et al.}}: A Sample Article Using IEEEtran.cls for IEEE Journals}
%
%\IEEEpubid{0000--0000/00\$00.00~\copyright~2021 IEEE}
% Remember, if you use this you must call \IEEEpubidadjcol in the second
% column for its text to clear the IEEEpubid mark.

\maketitle

\begin{abstract}
	Autonomous navigation in dense and highly dynamic environments requires both physically feasible control and low-latency replanning. Optimization-based methods such as Model Predictive Control (MPC) explicitly handle robot kinematics and safety constraints, but repeated nonlinear optimization can limit real-time responsiveness. Deterministic behavior-cloning policies enable efficient inference but may fail to represent multimodal avoidance behaviors, whereas diffusion policies capture multimodality at the cost of time-consuming iterative denoising. We propose PIER-Flow (Physics-Informed Efficient Rectified Flow), a lightweight navigation policy for mobile robots. By distilling an MPC expert into a continuous-time Ordinary Differential Equation (ODE), PIER-Flow achieves single-step action generation through parallel latent sampling and lightweight feasibility selection. We introduce a physics-informed training objective to enforce kinematic consistency, paired with an asynchronous action chunking architecture for robust sim-to-real deployment. Extensive simulations demonstrate that PIER-Flow achieves a 98.85\% success rate and zero collisions, with an average inference of $\sim$1.29 ms, which accelerates planning by 37.2$\times$ compared to MPC and over 800$\times$ against standard diffusion models. Crucially, real-world deployment on a resource-constrained edge computer further achieves an approximately stable inference latency of $\sim$5.3 ms, avoiding the latency spikes and freezing events observed with planning baselines.
\end{abstract}


\section{Introduction}
Navigating a mobile robot safely through dense and dynamic environments remains a fundamental challenge in robotics \cite{kretzschmar2016socially, fox1997dynamic, nahavandi2025comprehensive}. Model Predictive Control (MPC) addresses this problem by optimizing a finite-horizon trajectory subject to robot kinematics and collision constraints, providing valuable look-ahead capability in complex interactions \cite{borrelli2017predictive, brito2021where, brito2019model}. However, repeatedly solving a nonlinear program becomes increasingly expensive as the prediction horizon and obstacle count grow, while sensitivity to initialization may introduce variable convergence time and delayed control \cite{cleac2022algames}. Control Barrier Function (CBF) quadratic programs provide a computationally lighter alternative by minimally modifying a nominal command to enforce forward-invariance conditions \cite{ames2019control, morton2025oscbf}. Nevertheless, conventional CBF safety filters can be locally myopic and conservative, potentially causing excessive deceleration, deadlocks, and freezing-like behaviors in cluttered environments \cite{keyumarsi2025circulation, trautman2010unfreezing}.

Behavior Cloning (BC) provides an efficient alternative by learning a direct observation-to-action policy from expert demonstrations through supervised learning \cite{florence2022implicit, mandlekar2021matters}.  Action-sequence prediction and action chunking further improve temporal consistency and reduce error accumulation \cite{zhao2023learning}. However, explicit BC policies trained with pointwise regression losses typically learn deterministic and unimodal mappings. In ambiguous situations where several maneuvers are valid, such policies may average incompatible action modes \cite{shafiullah2022behavior}. Moreover, closed-loop execution exposes the policy to observations outside the expert distribution, causing covariate shift, compounding errors, and limited generalization to unseen interactions \cite{ross2011reduction}. 

These limitations have motivated generative policies that explicitly model conditional action distributions. Diffuser formulates trajectory planning as iterative denoising \cite{janner2022planning}, while Diffusion Policy generates observation-conditioned action sequences and captures multimodal robot behaviors \cite{chi2025diffusion}. Conditional diffusion models have also shown strong performance in mobile navigation \cite{sridhar2024nomad, yu2024ldp}, and recent work has learned diffusion controllers from optimization-based experts \cite{marquez2025diffusion}. However, diffusion inference requires multiple sequential neural-network evaluations. Physics-aware variants further increase this burden: SafeDiffuser introduces CBF-based guidance during reverse diffusion \cite{wang2023safediffuser}, whereas M2Diffuser applies differentiable optimization during denoising \cite{yan2025m2diffuser}. Although these methods improve safety and executability, iterative sampling and repeated correction conflict with the high-frequency, low-jitter requirements of edge-deployed chassis control.

Flow Matching provides promising alternatives for reducing generative inference cost \cite{ding2024consistency, lipman2022flow}. In particular, Rectified Flow (RF) learns an approximately straight transport path from noise to data, allowing accurate generation with fewer integration steps \cite{liu2022flow}. FlowMP introduces second-order motion information to improve trajectory smoothness and executability \cite{nguyen2025flowmp}. However, standard flow objectives primarily match demonstration distributions and do not inherently guarantee consistency with robot kinematics or safety constraints. Existing constraint-aware approaches commonly rely on test-time barrier correction, numerical integration, or sampling-based MPC refinement \cite{mizuta2025unified}. Such online procedures may offset the efficiency advantages of flow-based generation on resource-constrained hardware.

Motivated by this gap, we propose Physics-Informed Efficient Rectified Flow (PIER-Flow), a lightweight generative navigation framework that integrates finite-horizon decision making, multimodal behavior generation, real-time inference, and safety-aware execution. A differentiable robot kinematic rollout is embedded into the flow-matching objective, transferring kinematic regularization from online correction to offline training. At inference, a single flow step generates multiple action-chunk candidates in parallel, while a lightweight feasibility selector evaluates their imminent execution segments against current obstacle observations. This complementary design improves kinematic and collision feasibility without gradient-based projection or online trajectory optimization. Together with a lightweight LiDAR adapter and asynchronous action chunking, PIER-Flow supports continuous high-frequency control on resource-constrained edge platforms.

Our primary contributions are threefold:
\begin{enumerate}
	\item We propose PIER-Flow, a lightweight generative navigation framework that unifies finite-horizon action generation, multimodal decision making, real-time inference, and safety-aware execution for mobile robots. By generating parallel action-chunk candidates through a single Rectified Flow step, it preserves look-ahead behavior while avoiding deterministic mode averaging and iterative diffusion sampling.
	\item We introduce a physics-informed training objective based on differentiable kinematic rollout to improve kinematic consistency without online computational overhead. A lightweight feasibility selector further evaluates the imminent execution horizon and rejects potentially unsafe candidates without iterative test-time optimization.
	\item We design a complete edge-oriented deployment architecture and conduct extensive simulation and physical evaluations. PIER-Flow achieves a 98.85\% success rate with approximately $\sim$1.29 ms mean latency in simulation and maintains an approximately $\sim$5.3 ms inference latency on a computationally limited edge computer, substantially reducing the latency variation and freezing events observed with baselines.
\end{enumerate}

\section{Problem Formulation}
Let the system state be defined as $\mathbf{x}_t = [p_x, p_y, \theta, v_x, v_y, \omega]^\text{T} \in \mathbb{R}^6$, and the control input as $\mathbf{u}_t = [v_x^c, v_y^c, \omega^c]^\text{T} \in \mathbb{R}^3$. The discrete-time kinematics are $\mathbf{x}_{t+1} = f_{kin}(\mathbf{x}_t, \mathbf{u}_t)$. To navigate safely among $K$ dynamic obstacles, we formulate the task as a finite-horizon non-linear programming (NLP) at each time step $t$:
{\small
	\begin{subequations}\label{eq:mpc_problem}
		\begin{align}
			\min_{\mathbf{U}, \boldsymbol{\epsilon}} \quad & \sum_{k=0}^{N-1} \mathcal{J}(\mathbf{x}_k, \mathbf{u}_k) + \mathcal{J}_{T}(\mathbf{x}_N) + W_{col} \sum_{j=1}^K \sum_{k=0}^N \epsilon_{j,k}^2 \label{eq:mpc_obj} \\
			\text{s.t.} \quad & 
			\begin{aligned}[t]
				& \mathbf{x}_{k+1} = f_{kin}(\mathbf{x}_k, \mathbf{u}_k), \\
				& \mathbf{u}_{min} \le \mathbf{u}_k \le \mathbf{u}_{max}, \\
				& \| \mathbf{p}_k - \mathbf{p}_{obs, k}^{(j)} \|^2 + \epsilon_{j,k} \ge (R_{rob} + R_{obs}^{(j)} + d_{safe})^2, \\
				& \epsilon_{j,k} \ge 0, \quad \forall j \in \{1, \dots, K\}
			\end{aligned} \label{eq:mpc_constraints}
		\end{align}
	\end{subequations}
}where $\mathbf{U} = \{\mathbf{u}_0, \dots, \mathbf{u}_{N-1}\}$ is the optimized control sequence, $N$ is the horizon, $\mathbf{p}_k$ is the Cartesian position, and $\mathbf{p}_{obs, k}^{(j)}$ is the predicted position of obstacle $j$. Slack variables $\epsilon_{j,k}$ ensure recursive feasibility. While solving this NLP problem guarantees theoretical safety and optimal topological behavior, computing the exact solution online in complex multi-obstacle scenarios is computationally prohibitive. The $\mathcal{O}(N \times K)$ complexity frequently leads to severe latency spikes, highlighting the critical need for a reactive, constant-time approximation.

\section{Methodology}
\begin{figure*}[htbp]
	\centering
	\includegraphics[width=1.98\columnwidth]{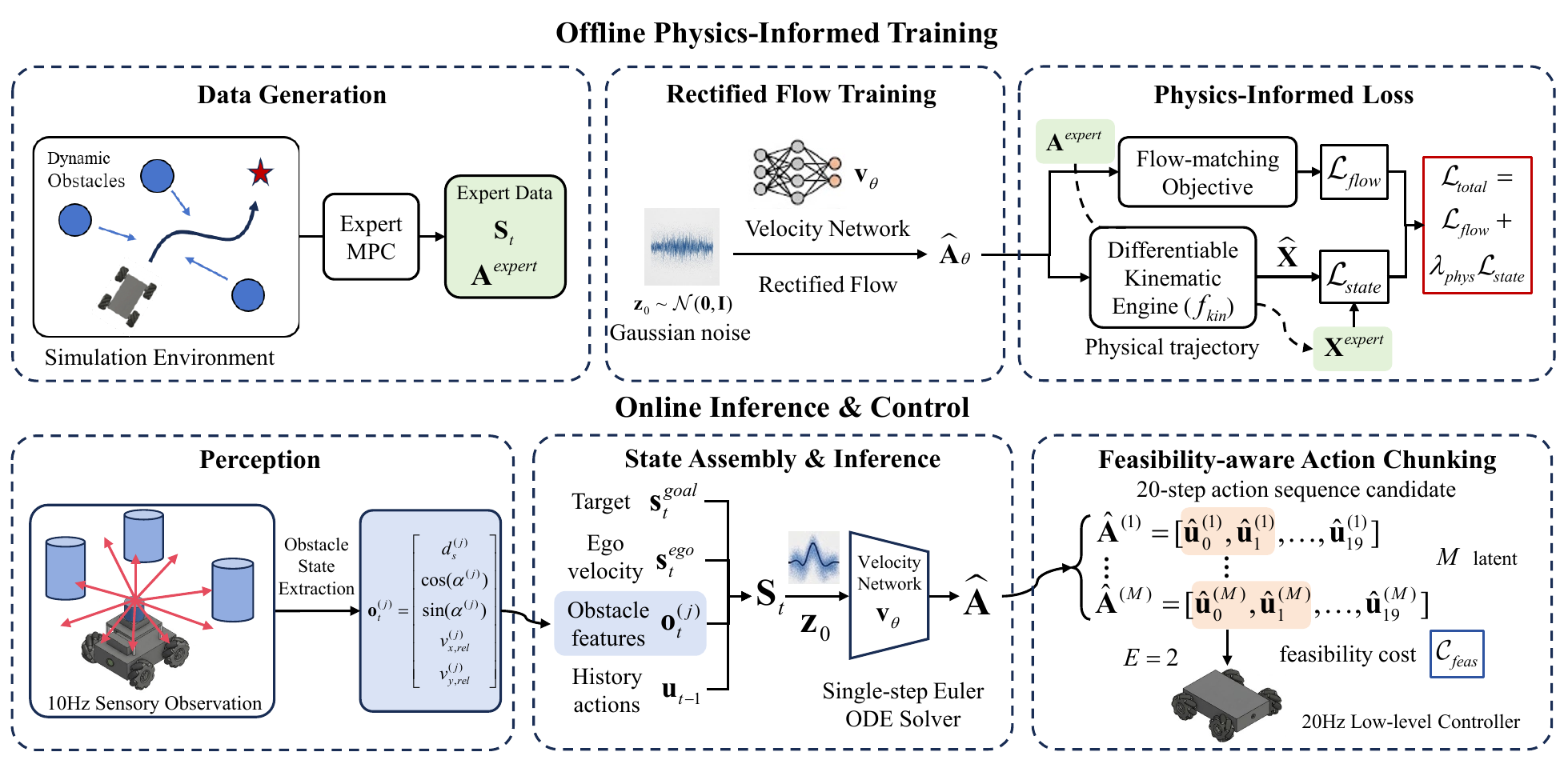}
	\caption{Overview of the proposed Physics-Informed Efficient Rectified Flow (PIER-Flow) navigation framework. \textbf{Offline physics-informed training:} An MPC expert generates state-action sequences in simulated randomized dynamic environments. Rectified Flow constructs a straight transport path from Gaussian noise $\mathbf z_0$ to the scaled expert action chunk $\mathbf z_1$. The training objective combines an action-space flow-matching loss $\mathcal L_{flow}$ with a trajectory-level state loss $\mathcal L_{state}$, where both the reconstructed action chunk and the expert chunk are propagated through the same differentiable kinematic rollout. \textbf{Online inference and control:} The current observation includes local obstacle features and the previously executed action. A single Euler step generates $M$ action-chunk candidates from a fixed latent bank in parallel. A lightweight feasibility selector evaluates the imminent $E$ execution steps, after which the selected commands are dispatched through an asynchronous command buffer to the high-frequency chassis controller.}
	\label{fig:system_architecture}
\end{figure*}

As illustrated in Fig.~\ref{fig:system_architecture}, PIER-Flow consists of two stages: offline physics-informed training and online inference and control. An MPC expert first generates state-action chunks, from which PIER-Flow learns a multimodal action distribution regularized by differentiable kinematic rollout. Single-step parallel generation and lightweight feasibility selection provide action chunks to the high-frequency chassis controller.

% [PLACEHOLDER: Insert "fig_system_architecture.png" here]

\subsection{State Representation and Expert Data Generation}
We formulate local navigation as the conditional generation of a horizon-$H$ body-frame velocity chunk. Given an observation $\mathbf S_t\in\mathbb R^{24}$, the policy generates
{\small
\begin{equation}
	\hat{\mathbf{A}}_t =
	[\hat{\mathbf{u}}_{t,0}, \hat{\mathbf{u}}_{t,1}, \ldots, \hat{\mathbf{u}}_{t,H-1}]
	\in \mathbb{R}^{3H},
\end{equation}
}where $\hat{\mathbf{u}}_{t,h}=[v_x,v_y,\omega]^\text{T}$ is the body-frame chassis command at prediction step $h$. We use
$H=20$ and $\Delta t=0.05$~s, resulting in a 1.0~s predictive horizon.

The egocentric observation is
{\small
\begin{equation}
	\mathbf{S}_t = \big[ (\mathbf{s}^{goal}_t)^\text{T}, (\mathbf{s}^{ego}_t)^\text{T}, (\mathbf{o}^{(1)}_t)^\text{T}, \dots, (\mathbf{o}^{(K)}_t)^\text{T}, \mathbf{u}_{t-1}^\text{T} \big]^\text{T}
\end{equation}
}

Specifically, $\mathbf{s}^{goal}_t=[d_g,\cos\alpha_g,\sin\alpha_g]^\text{T} \in \mathbb{R}^3$ represents the polar coordinates of the relative goal, and $\mathbf{s}^{ego}_t=[v_x,v_y,\omega]^\text{T} \in \mathbb{R}^3$ captures the ego-velocity. We track the $K=3$ most threatening obstacles, encoding each as $\mathbf{o}^{(j)}_t = [d_{s}^{(j)}, \cos(\alpha^{(j)}), \sin(\alpha^{(j)}), v_{x, rel}^{(j)}, v_{y, rel}^{(j)}]^\text{T} \in \mathbb{R}^5$. Dummy values are padded when fewer than three obstacles are detected. 

\textbf{Short-term Intention Memory.}
Symmetric scenes may admit distinct valid behaviors, such as passing an obstacle from either side. Small observation perturbations can therefore cause a generative policy to switch repeatedly between behavioral modes. To provide short-term intention memory and reduce such mode switching, we augment the current observation with the previously executed command
$\mathbf u_{t-1}\in\mathbb R^3$. 

\textbf{Expert Data Collection via MPC.} To collect high-quality expert demonstrations, we employ an MPC solver \cite{borrelli2017predictive} to resolve the optimal control problem defined in Eq.~(\ref{eq:mpc_problem}) across randomized dynamic environments. To improve gradient flow during neural network training, all velocity actions are normalized via an amplification scalar $\gamma_{scale}$. Only collision-free trajectories are recorded, yielding a dataset of optimal topological decisions where the expert action chunk is defined as $\mathbf{A}^{expert} = [\mathbf{u}^{expert}_0, \dots, \mathbf{u}^{expert}_{H-1}] \in \mathbb{R}^{3H}$.

\subsection{Lightweight Physics-Informed Rectified Flow Policy}
We formulate local navigation as the conditional generation of expert velocity chunks. Let $\mathbf{z}_1\sim \gamma_{scale} \cdot {{p}_{expert}}({{\mathbf{A}}^{expert}}|{{\mathbf{S}}_{t}})$ denote the scaled expert chunk, and let $\mathbf{z}_0\sim\mathcal{N}(\mathbf{0},\mathbf{I})$ be the base noise. Rectified Flow constructs a straight transport path between the noise and the expert action:
{\small
\begin{equation}
	\mathbf{z}_{\tau}=(1-\tau)\mathbf{z}_0+\tau\mathbf{z}_1,
	\quad \tau\sim\mathcal{U}(0,1),
\end{equation}
}whose target velocity is $\mathbf{v}^{*}=\mathbf{z}_1-\mathbf{z}_0$. We train a conditional velocity field $\mathbf{v}_{\theta}(\mathbf{z}_{\tau},\tau,\mathbf{S}_t)$ using the flow-matching objective:
{\small
\begin{equation}
	\mathcal{L}_{flow}
	=
	\mathbb{E}
	\left[
	\left\|
	\mathbf{v}_{\theta}(\mathbf{z}_{\tau},\tau,\mathbf{S}_t)
	-(\mathbf{z}_1-\mathbf{z}_0)
	\right\|_2^2
	\right].
\end{equation}
}

The velocity network $\mathbf{v}_\theta$ is a compact multi-layer perceptron (MLP) conditioned on both the noisy action and the current state. The flow time $\tau$ is embedded via a 64-dimensional Gaussian Fourier projection. The network input is $[\mathbf{z}_{\tau},\phi(\tau),\mathbf{S}_t]\in\mathbb{R}^{60+64+24}$, and the main MLP contains three 1024-unit hidden layers with SiLU activations. The 60-dimensional output predicts the full $H=20$ action chunk. This architecture contains approximately 2.32M parameters, ensuring suitability for onboard edge inference.

To improve physical consistency, we augment the standard flow objective with a differentiable kinematic unrolling loss. For each sampled point on the PIER-Flow transport path, the model implies a predicted endpoint:
{\small
\begin{equation}
	\hat{\mathbf{z}}_1
	=
	\mathbf{z}_{\tau}
	+(1-\tau)
	\mathbf{v}_{\theta}(\mathbf{z}_{\tau},\tau,\mathbf{S}_t),
	\quad
	\hat{\mathbf{A}}_{\theta}=\hat{\mathbf{z}}_1/\gamma_{scale}.
\end{equation}
}

After reshaping $\hat{\mathbf{A}}_{\theta}$ into horizon-$H$ commands $[v_x,v_y,\omega]^\text{T}$, we unroll the omnidirectional kinematics from $(x_0,y_0,\theta_0)=(0,0,0)$:
{\small
\begin{subequations}\label{eq:robot_kinematics}
\begin{align}
	x_{h+1} &= x_h + (v_{x,h}\cos\theta_h - v_{y,h}\sin\theta_h)\Delta t,\\
	y_{h+1} &= y_h + (v_{x,h}\sin\theta_h + v_{y,h}\cos\theta_h)\Delta t,\\
	\theta_{h+1} &= \theta_h + \omega_h\Delta t.
\end{align}
\end{subequations}
}

Let $\mathbf{x}_h = [\hat{x}_h,\hat{y}_h,\hat{\theta}_h]^\text{T}$ and $\mathbf{x}^{expert}_h=[x^{expert}_h,y^{expert}_h,\theta^{expert}_h]^\text{T}$. Starting from the same initial pose, Eq.~(\ref{eq:robot_kinematics}) is recursively applied to the predicted commands in $\hat{\mathbf{A}}_{\theta}$ and the expert commands in $\mathbf{A}^{expert}$, respectively. This yields $\hat{\mathbf{X}}_{1:H}=[\hat{\mathbf{x}}_1,\ldots,\hat{\mathbf{x}}_H]$ and $\mathbf{X}^{expert}_{1:H}=[\mathbf{x}^{expert}_1,\ldots,\mathbf{x}^{expert}_H]$. The physics-informed state loss is calculated as:
{\small
\begin{equation}
	\mathcal{L}_{state}
	=
	\mathbb{E}
	\left[
	\frac{1}{H}
	\sum_{h=1}^{H}
	\left\|
	\hat{\mathbf{x}}_h-\mathbf{x}^{expert}_h
	\right\|_2^2
	\right]
\end{equation}
}
{\small
\begin{equation}
	\mathcal{L}_{total}
	=
	\mathcal{L}_{flow}
	+\lambda_{phys}\mathcal{L}_{state}
\end{equation}
}where $\lambda_{phys}$ is an empirically selected weighting coefficient. The state loss acts as an inductive bias that penalizes the accumulated trajectory-level deviation between the predicted and expert actions under the robot kinematics.

\subsection{Feasibility-aware Action Chunking}
A critical challenge in deploying generative navigation policies on mobile platforms is the frequency mismatch between the planning loop and the low-level chassis control. Let $f_{plan}$ denote the frequency of sensory perception and neural inference, and $f_{ctrl}$ denote the high-frequency requirement of the motor controller. To address this mismatch without reintroducing online trajectory optimization, we combine parallel candidate generation, lightweight feasibility selection, and action chunking.

\textbf{Parallel Latent Candidate Generation.} Instead of drawing a single random sample or relying purely on the expected mean, we construct a parallel batch of $M$ latent candidates during each replanning cycle: $\mathcal{Z}_0 = [\mathbf{z}_0^{(1)}, \dots, \mathbf{z}_0^{(M)}]$. To aggressively exploit the dominant deterministic mode while maintaining exploration, the first candidate is anchored at the noise mean ($\mathbf{z}_0^{(1)} = \mathbf{0}$). The remaining $M-1$ candidates utilize fixed random seeds ($\mathbf{z}_0^{(m)} \sim \mathcal{N}(\mathbf{0}, \mathbf{I})$ for $m > 1$) to provide topological diversity while preserving strict temporal consistency across consecutive sensory frames. 

By leveraging parallel computation, the PIER-Flow ODE $d\mathbf{z}_{\tau}/d\tau=\mathbf{v}_{\theta}(\mathbf{z}_{\tau},\tau,\mathbf{S}_t)$ is solved using a fast, single-step Euler integration over the batch $\mathcal{Z}_0$. This simultaneously yields $M$ candidate trajectory chunks, $\{\hat{\mathbf{A}}^{(1)}_t, \dots, \hat{\mathbf{A}}^{(M)}_t\}$, each covering a horizon of $H$ steps.

\textbf{Lightweight Feasibility Selector.} To select an executable chunk without invoking an online MPC solver, we introduce a lightweight feasibility selector. The execution engine calculates the required execution steps $E=f_{ctrl}/f_{plan}$. For each candidate chunk, only the imminent $E$ execution steps are evaluated, since these commands will be executed before the next replanning cycle. The feasibility cost is defined as:
{\small
\begin{equation}
	\mathcal{C}_{feas}(\hat{\mathbf{A}}) = \sum_{e=0}^{E-1} \Big( c_{vel}(\hat{\mathbf{u}}_{t,e}) + c_{yaw}(\hat{\mathbf{u}}_{t,e}) + c_{obs}(\hat{\mathbf{x}}_{t,e}, \mathbf{O}_t) \Big)
\end{equation}
}where $c_{vel}$ and $c_{yaw}$ heavily penalize violations of the predefined chassis velocity and angular limits, and $c_{obs}$ penalizes intrusion into the inflated safety margins of perceived dynamic obstacles ($\mathbf{O}_t$). 

The selected action chunk is then obtained by $\hat{\mathbf{A}}^{*}_t = \arg\min_{m \in \{1, \dots, M\}} \mathcal{C}_{feas}(\hat{\mathbf{A}}^{(m)}_t)$. This selector screens potentially unsafe candidates over the imminent execution horizon.

\textbf{Asynchronous Action Chunking.} Finally, the low-level controller sequentially executes the first $E$ velocity commands of $\hat{\mathbf{A}}^{*}_t$ at frequency $f_{ctrl}$. At the next planning cycle, a new action chunk is generated and selected based on the updated observation.
\begin{figure}[tbp]
	\centering
	\includegraphics[width=0.96\columnwidth]{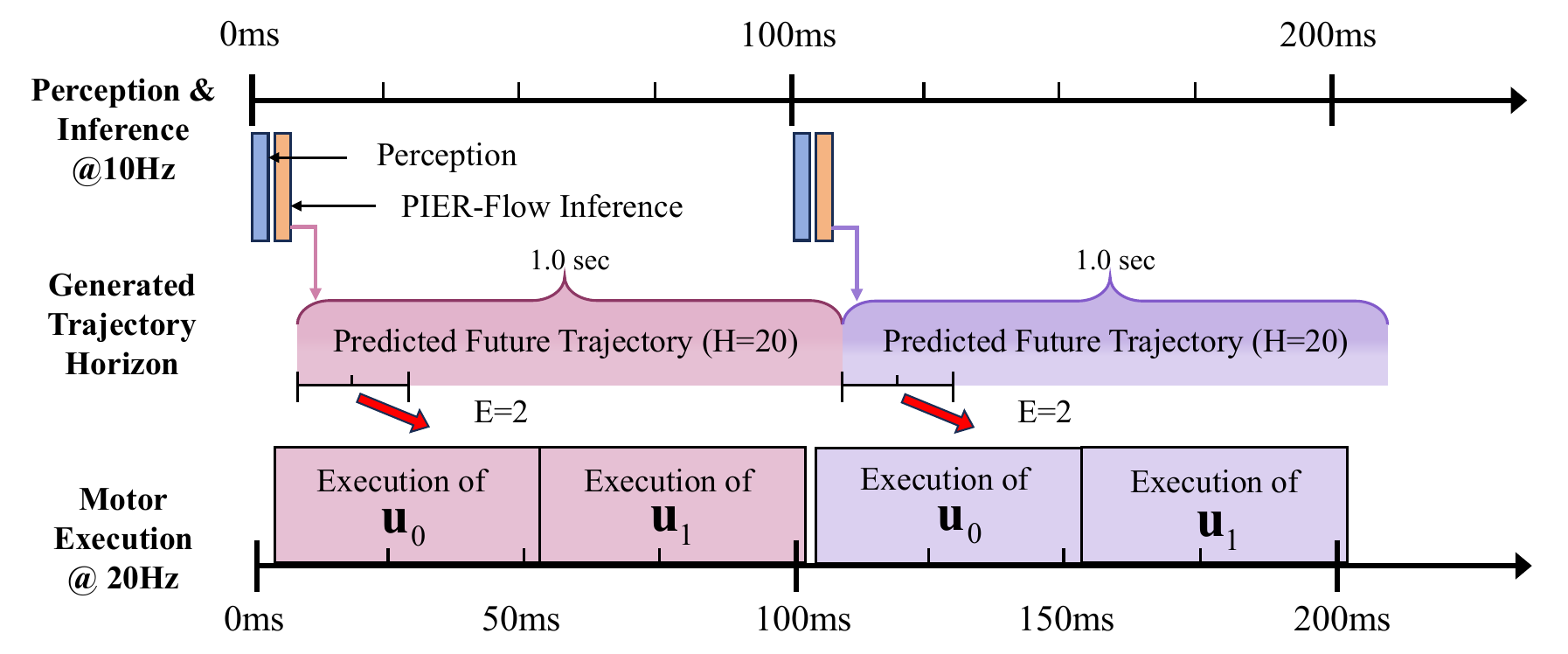}
	\caption{Asynchronous timing diagram illustrating the action chunking strategy. The perception and PIER-Flow inference loop operates at 10 Hz, generating a 20-step predictive trajectory horizon ($H=20$). The low-level chassis controller sequentially executes the first two velocity commands ($E=2$) at 20 Hz, ensuring continuous and smooth physical motion.}
	\label{fig:timing_diagram}
\end{figure}

For instance, illustrated in Fig.~\ref{fig:timing_diagram}, given an inference loop of 10 Hz and a motor control loop of 20 Hz, the execution ratio is $E=2$. The action chunking mechanism bridges the planning-control frequency gap by providing continuous high-frequency commands derived from the selected generative action chunk.

\section{Simulation Experiments}
To rigorously evaluate the proposed PIER-Flow navigation policy, we conduct extensive simulations in a highly dynamic multi-obstacle environment. We compare our approach against learning-based and optimization-based baselines.

\subsection{Expert Data Generation and Training Setup}
To generate the expert demonstrations, we employ the MPC controller in a custom mecanum-wheel robot simulator ($\Delta t=0.05$ s). Each training episode initializes the robot near the origin with a randomized position ($\pm 0.2$ m) and randomized initial body-frame velocities to expose the policy to nonzero transient states. The goal is fixed at $(3.0, 3.0)$ m with a constant heading of $\pi/4$. The environment features three circular dynamic obstacles placed along the start-goal corridor. These obstacles are assigned lateral offsets and move perpendicularly to the nominal path with random speeds up to $1.0$ m/s.

At every control step, the MPC solves a horizon-20 local planning problem with velocity bounds of $\pm 1.0$ m/s on $v_x$ and $v_y$, zero commanded yaw rate, and a safety margin of $0.2$ m around each obstacle. To improve policy robustness, occasional safe state perturbations are injected during data collection; perturbations that would place the robot inside an inflated obstacle boundary are rejected. Retaining only successful, collision-free MPC episodes yields 2000 expert rollouts, comprising exactly 166,599 supervised samples. These collected training scenarios are stored separately from the evaluation set.

During PIER-Flow training, expert actions are scaled by $\gamma_{scale}=10$ to improve gradient flow and unscaled post-inference. The dataset is split into 90\% training and 10\% validation subsets. The velocity network is trained using the Adam optimizer. The final PIER-Flow model utilizes the physics-informed loss empirically weighted at $\lambda_{phys}=10$.

\subsection{Comparative Analysis}
We benchmark our method against five baselines over 2000 randomized evaluation cases: 1) MPC: The expert non-linear predictive controller \cite{borrelli2017predictive}; 2) CBF: A reactive Control Barrier Function method \cite{morton2025oscbf}; 3) BC-MLP: A standard Behavior Cloning MLP trained via pure regression \cite{mandlekar2021matters}; 4) DDPM: A standard diffusion policy using 100 denoising steps \cite{chi2025diffusion}; 5) DDIM: An accelerated diffusion policy using 8 denoising steps \cite{chi2025diffusion}. To ensure a fair comparison, both learning-based baselines (BC-MLP, DDPM and DDIM) are trained on the exact same expert demonstration dataset as our proposed PIER-Flow policy, sharing identical state-action representations. For our PIER-Flow online inference, we set the number of parallel candidates to $M=2$ (the noise mean and one fixed-seed latent) to balance topological diversity with ultra-low computational overhead. The simulated benchmarking was conducted on a workstation equipped with an NVIDIA GeForce RTX 3090 GPU. The quantitative results are summarized in Table~\ref{tab:main_results}.

\begin{table*}[htbp]
	\caption{Quantitative Comparison on the Multi-Obstacle Dynamic Navigation Benchmark}
	\label{tab:main_results}
	\centering
	\begin{tabular}{l c c c c c c}
		\toprule
		\textbf{Method} & \textbf{Success (\%)} $\uparrow$ & \textbf{Collision (\%)} $\downarrow$ & \textbf{Timeout (\%)} $\downarrow$ & \textbf{Min Clearance (m)} $\uparrow$ & \textbf{Mean Latency (ms)} $\downarrow$ & \textbf{P95 Latency (ms)} $\downarrow$ \\
		\midrule
		MPC & 97.45 & \textbf{0.00} & 2.55 & 0.206 & 48.01 & 52.24 \\
		CBF & 87.15 & \textbf{0.00} & 12.85 & \textbf{0.309} & \textbf{0.18} & \textbf{0.18} \\
		BC-MLP & 98.20 & 0.30 & 1.50 & 0.202 & 0.54 & 0.55 \\
		DDPM & 97.80 & 1.35 & 0.85 & 0.189 & 1077.24 & 1081.10 \\
		DDIM & 97.85 & 1.40 & \textbf{0.75} & 0.189 & 84.73 & 85.41 \\
		\textbf{Ours} & \textbf{98.85} & \textbf{0.00} & 1.15 & 0.201 & 1.29 & 1.40 \\
		\bottomrule
	\end{tabular}
\end{table*}

% [PLACEHOLDER: Insert "fig_main_success_failure_latency.png" here]
\begin{figure}[htbp]
	\centering
	\includegraphics[width=0.98\columnwidth]{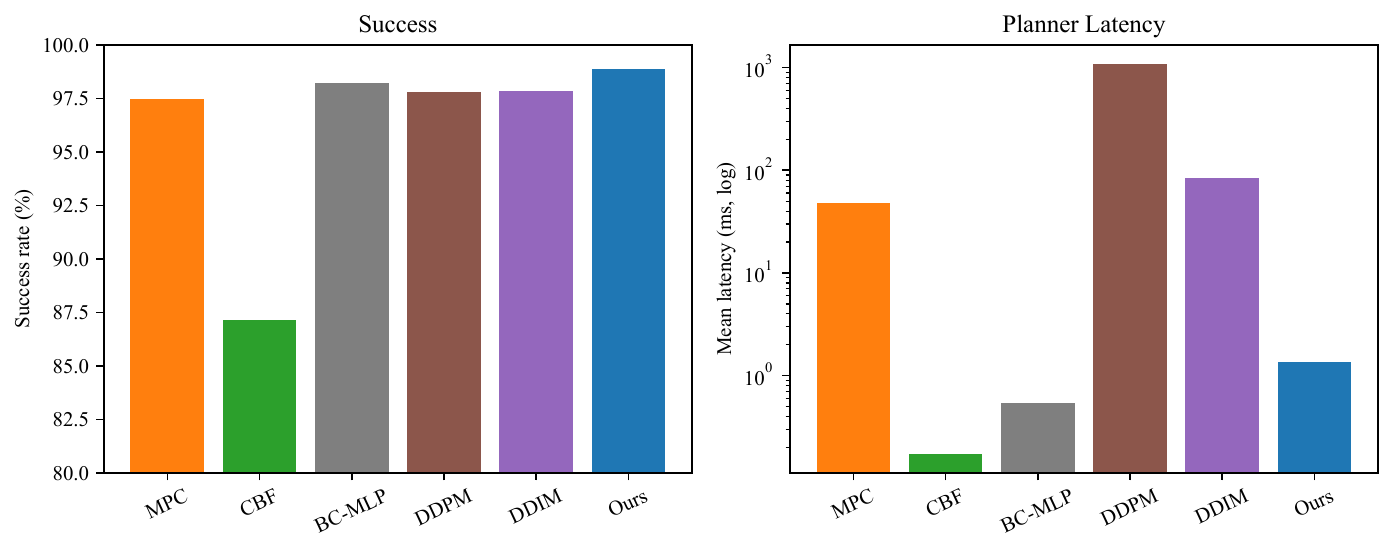}
	\caption{Comparison of success rates and inference latency (log-scale).}
	\label{fig:main_performance}
\end{figure}

As shown in Table~\ref{tab:main_results}, our proposed PIER-Flow achieves the highest success rate (98.85\%) and strictly maintains zero collisions, successfully distilling the safety guarantees of the expert MPC. While the purely regression-based BC-MLP achieves faster neural inference (0.54 ms), it incurs a 0.30\% collision rate. This highlights the inherent limitation of standard behavior cloning: it struggles to capture the multi-modal nature of expert demonstrations. DDIM and DDPM similarly exhibit higher collision rates in this benchmark, indicating potential vulnerabilities in highly cluttered scenarios. The purely reactive CBF achieves the absolute lowest inference time (0.18 ms) but suffers from a significant timeout rate (12.85\%). Lacking long-horizon foresight, the CBF aggressively decelerates to maintain strict safety invariants when confronted with dense dynamic obstacles, thereby triggering the classic ``freezing robot" problem.

Crucially, the proposed PIER-Flow policy resolves the computation bottleneck inherent to robust generative navigation models. As illustrated by the quantitative results, PIER-Flow completes batch inference and feasibility selection in merely 1.29 ms on average, running approximately 37.2$\times$ faster than the MPC solver, 65.7$\times$ faster than DDIM, and over 800$\times$ faster than standard DDPM. While marginally slower than the naive BC-MLP, this near-millisecond execution provides collision safety and topological diversity, making the policy highly suitable for integration into high-frequency embedded real-time control loops.

\subsection{Ablation Studies}

\subsubsection{Effect of Integration Sampling Steps}
\begin{figure}[htbp]
	\centering
	\includegraphics[width=0.96\columnwidth]{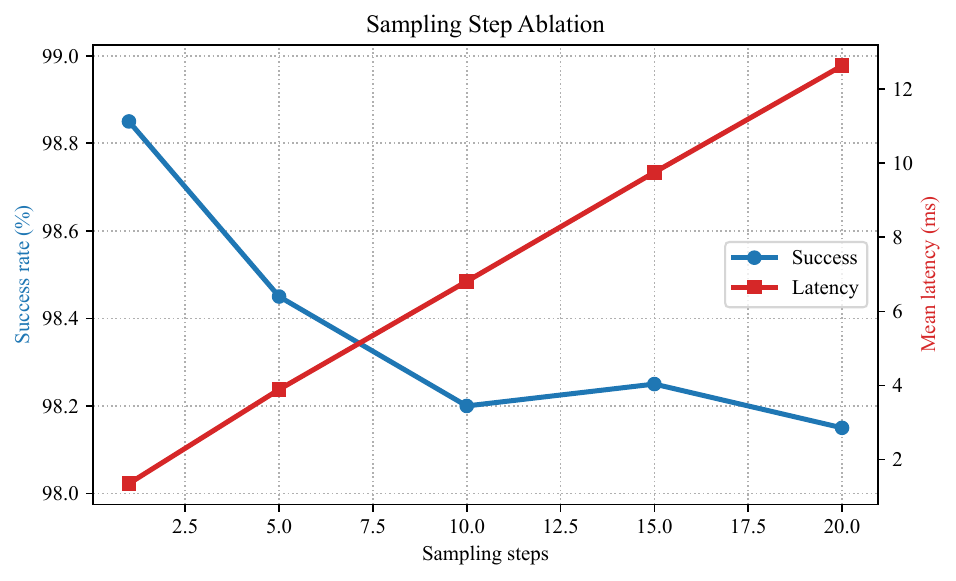}
	\caption{Impact of integration steps on navigation success rate and inference latency.}
	\label{fig:step_ablation}
\end{figure}

Generative models typically face a fundamental trade-off between sample quality and inference speed, dictated by the number of integration steps. We analyze this trade-off for our trained PIER-Flow policy in Fig.~\ref{fig:step_ablation}.

%\begin{table}[htbp]
%	\caption{Ablation on Rectified Flow Integration Steps}
%	\label{tab:ablation_steps}
%	\centering
%	\begin{tabular}{c c c c}
%		\toprule
%		\textbf{Steps} & \textbf{Success (\%)} $\uparrow$ & \textbf{Collision (\%)} $\downarrow$ & \textbf{Mean Latency (ms)} $\downarrow$ \\
%		\midrule
%		\textbf{1} & \textbf{98.60} & \textbf{0.00} & \textbf{0.91} \\
%		5 & 98.15 & 0.25 & 3.29 \\
%		10 & 97.85 & 0.30 & 6.09 \\
%		15 & 97.50 & 0.40 & 9.50 \\
%		20 & 97.45 & 0.45 & 12.98 \\
%		\bottomrule
%	\end{tabular}
%\end{table}

% [PLACEHOLDER: Insert "fig_rf_sampling_ablation.png" here]

Counter-intuitively, increasing the number of solver steps does not yield better navigation performance. The single-step solver not only achieves the fastest inference but also strictly dominates in success rate and safety. This strongly validates our design choice: the straight-line probability paths learned by the flow-matching objective enable the policy to jump directly from the noise distribution to the expert action distribution in a single robust step.

\subsubsection{Impact of the Physics-Informed Loss}
To evaluate the contribution of the physics-guided training objective, we compare the full PIER-Flow pipeline against an action-only RF baseline trained exclusively on regression loss.

\begin{table}[htbp]
	\caption{Ablation on Physics-Informed Training}
	\label{tab:ablation_physics}
	\centering
	\begin{tabular}{l c c c}
		\toprule
		\textbf{Variant} & \textbf{Success (\%)} & \textbf{Collision (\%)} & \textbf{Timeout (\%)} \\
		\midrule
		Action-only RF & 98.10 & 0.10 & 1.80 \\
		\textbf{Ours} & \textbf{98.85} & 0.00 & \textbf{1.15} \\
		\bottomrule
	\end{tabular}
\end{table}

While the action-only baseline already demonstrates strong imitation capabilities, integrating the differentiable kinematic constraints further improves the success rate and reduces collisions and timeouts (Table~\ref{tab:ablation_physics}). More importantly, rather than pursuing a dramatic numerical gain in pure simulation, the primary role of the physics-informed loss is to serve as an inductive bias. It strictly penalizes dynamically infeasible trajectory predictions during training, yielding actions that respect chassis mobility limits. This kinematic consistency drastically minimizes the sim-to-real gap, ensuring the policy's safe deployability on physical mobile robots.

\section{Hardware Validation and Real-World Experiments}
To validate the sim-to-real transferability and the real-time execution capabilities of the proposed PIER-Flow policy, we deployed our algorithm on a physical omnidirectional mobile robot.

\subsection{Hardware Setup and System Implementation}
The physical experiments were conducted on a ROSMASTER X3 mobile robot equipped with mecanum wheels. For environmental perception, the robot utilizes a high-frequency RPLIDAR-S2, providing continuous 2D point cloud scans. All computations, including perception, state estimation, and neural network inference, are executed entirely onboard using an NVIDIA Jetson Orin Nano edge computer running the ROS2 framework.

\textbf{Real-time Perception Pipeline.}
To bridge the reality gap between simulation states and physical LiDAR data, we implemented a lightweight onboard tracking module. Raw LiDAR scans are truncated to a valid range of $0.25$ m to $3.0$ m and transformed into the robot's odometry frame. We apply a Non-Maximum Suppression (NMS) \cite{neubeck2006efficient} clustering algorithm with a spatial threshold of $0.5$ m to isolate distinct physical entities. To estimate the velocity of dynamic obstacles, a temporal matching filter is applied across consecutive frames, utilizing an exponential moving average to smooth instantaneous velocity vectors. 

\textbf{Action Chunking Execution.}
Operating generative models on edge hardware demands rigorous timing management. Our PIER-Flow model infers a future trajectory of 20 velocity commands ($\Delta t = 0.05$ s). To adhere to the real-time constraints, we set the parallel generation batch size to $M=2$. We employ the action chunking strategy, in which the low-level chassis controller sequentially executes the first $E=2$ steps of the predicted chunk at a control rate of 20 Hz, thereby masking the neural network's inference latency before the next LiDAR frame triggers the replanning cycle.

\subsection{Evaluation Scenarios and Baselines}
We established four distinct physical testing scenarios to progressively stress-test the navigation algorithms: 1) \textbf{Single Static}, 2) \textbf{Multi Static}, 3) \textbf{Single Dynamic}, and 4) \textbf{Multi Dynamic}. 

We compared our PIER-Flow method against three baselines: the expert MPC  \cite{borrelli2017predictive}, a reactive CBF \cite{morton2025oscbf}, and a DDIM \cite{chi2025diffusion}. The standard DDPM \cite{chi2025diffusion} was explicitly excluded from hardware evaluation, as its prolonged multi-step inference fundamentally violates the control loop constraints of physical mobile platforms. Furthermore, BC-MLP \cite{mandlekar2021matters} was also excluded from physical deployment because it struggles to capture the multimodal characteristics of expert demonstrations, making it ill-suited for complex obstacle avoidance scenarios.

\subsection{Real-World Performance and Behavioral Analysis}

\begin{figure}[htbp]
	\centering
	\includegraphics[width=0.96\columnwidth]{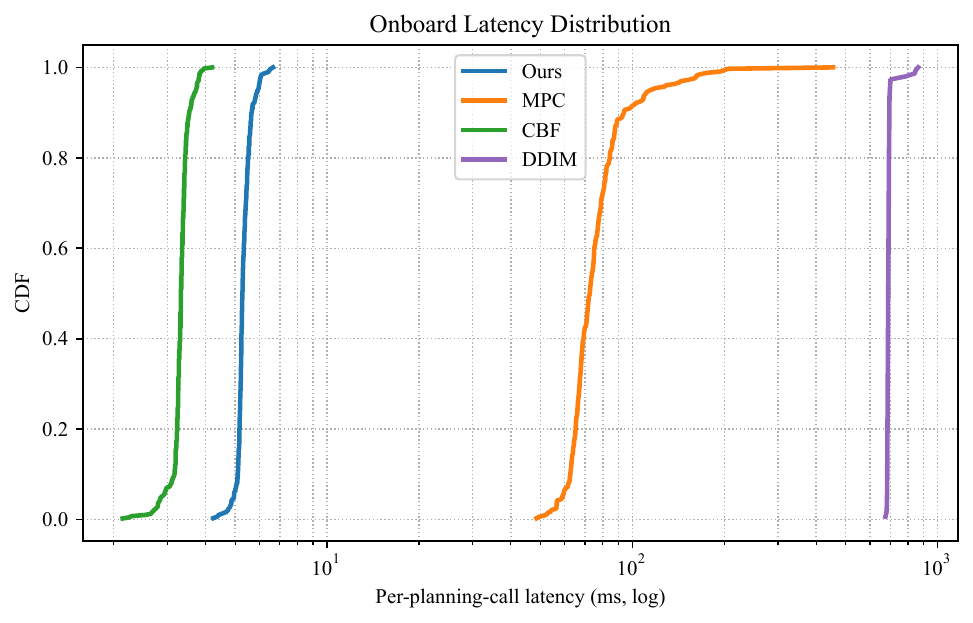}
	\caption{Per-step Latency CDF recorded during physical deployment. Our PIER-Flow policy exhibits a near-vertical, deterministic timing profile.}
	\label{fig:hw_latency_cdf}
\end{figure}

\begin{figure}[htbp]
	\centering
	\includegraphics[width=0.9\columnwidth]{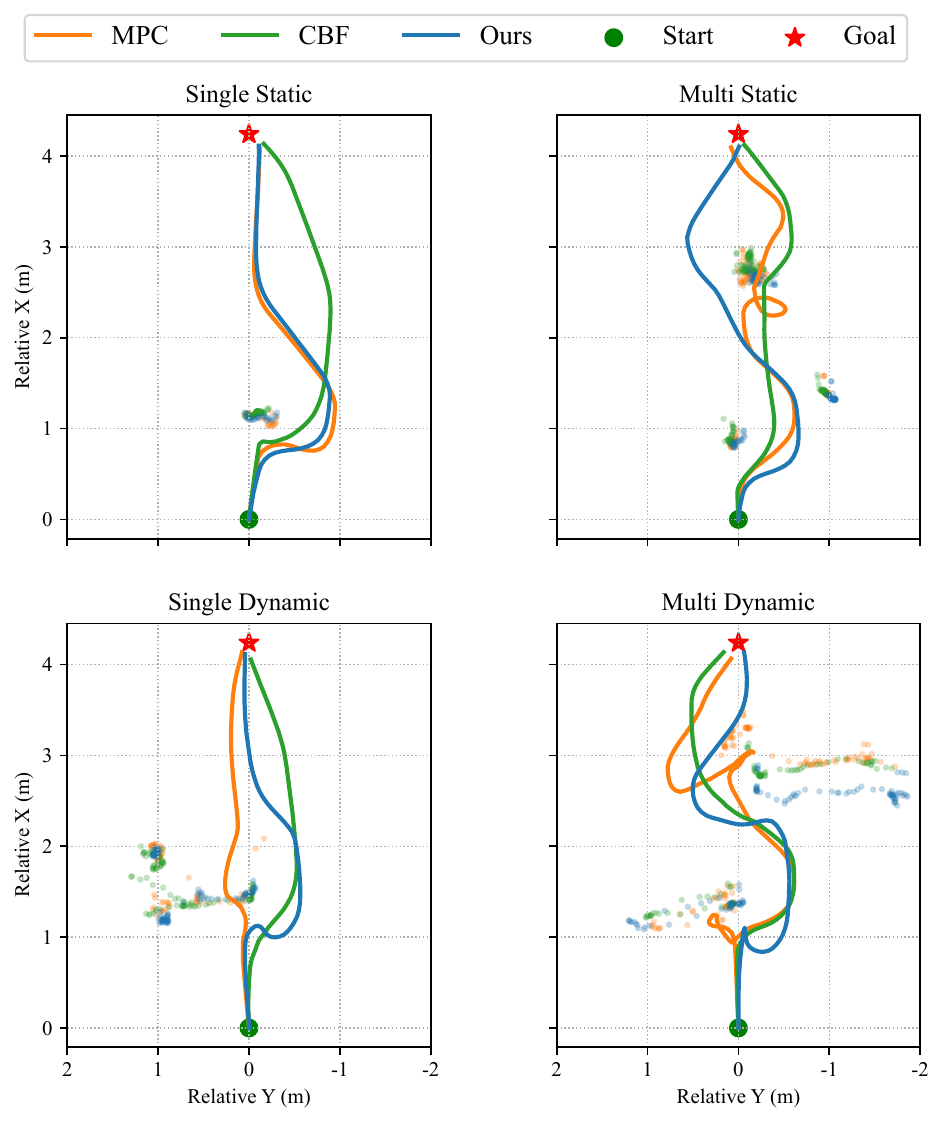}
	\caption{Trajectory visualization for Ours, MPC, and CBF across the four evaluation scenarios.}
	\label{fig:hw_trajectory}
\end{figure}

The quantitative performance across all real-world scenarios is summarized in Table~\ref{tab:real_world_results}, while the latency distribution and trajectory behaviors are visualized in Figs.~\ref{fig:hw_latency_cdf} and \ref{fig:hw_trajectory}. Overall, PIER-Flow succeeds in all four scenarios and maintains a stable onboard inference latency of 5.3~ms, with the maximum latency bounded by 6.65~ms. In contrast, the baselines expose different failure modes: MPC suffers from latency spikes, CBF exhibits conservative freezing-like behavior, and DDIM is too slow for real-time physical deployment.

\begin{table*}[htbp]
	\caption{Overall Real-Robot Experimental Results on the ROSMASTER X3}
	\label{tab:real_world_results}
	\centering
	\resizebox{\textwidth}{!}{%
		\begin{tabular}{l l l c c c c}
			\toprule
			\textbf{Scenario} & \textbf{Method} & \textbf{Outcome} & \textbf{Time (s)} $\downarrow$ & \textbf{Path (m)} & \textbf{Mean Latency (ms)} $\downarrow$ & \textbf{Max Latency (ms)} $\downarrow$ \\
			\midrule
			\multirow{4}{*}{Single Static} 
			& MPC & Success & 7.02 & 5.08 & 69.22 & 108.78 \\
			& CBF & Success & 12.27 & 4.73 & \textbf{3.27} & \textbf{3.75} \\
			& DDIM & Fail & - & - & 692.10 & 863.00 \\
			& \textbf{Ours} & \textbf{Success} & \textbf{6.57} & 4.83 & 5.32 & 6.65 \\
			\midrule
			\multirow{4}{*}{Multi Static} 
			& MPC & Collision & 9.22 & 5.71 & 80.98 & 194.10 \\
			& CBF & Success & 9.78 & 4.51 & \textbf{3.36} & \textbf{3.82} \\
			& DDIM & Fail & - & - & 691.51 & 784.20 \\
			& \textbf{Ours} & \textbf{Success} & \textbf{7.22} & 4.98 & 5.33 & 6.01 \\
			\midrule
			\multirow{4}{*}{Single Dynamic} 			
			& MPC & Collision & 6.18 & 4.26 & 74.45 & 162.03 \\
			& CBF & Success & 7.20 & 4.27 & \textbf{3.27} & \textbf{3.94} \\
			& DDIM & Fail & - & - & 693.00 & 841.50 \\
			& \textbf{Ours} & \textbf{Success} & \textbf{6.71} & 4.83 & 5.35 & 6.46 \\
			\midrule
			\multirow{4}{*}{Multi Dynamic} 			
			& MPC & Collision & 12.88 & 7.33 & 85.10 & 454.53 \\
			& CBF & Success & 8.98 & 5.04 & \textbf{3.39} & \textbf{4.20} \\
			& DDIM & Fail & - & - & 694.94 & 846.50 \\
			& \textbf{Ours} & \textbf{Success} & \textbf{8.46} & 5.87 & 5.36 & 6.53 \\
			\bottomrule
		\end{tabular}%
	}
\end{table*}

\textbf{Latency Stability and Real-world Failures.}
The onboard latency distributions explain the performance gap between simulation and physical deployment. DDIM consistently requires nearly 700~ms per inference step on the edge computer, making the robot unable to react to dynamic changes in time. As a result, DDIM fails in all real-world scenarios and must be manually terminated. Although MPC serves as a strong optimization-based expert in simulation, its nonlinear solver exhibits a long-tail latency distribution on onboard hardware. In the most complex Multi Dynamic scenario, the maximum MPC latency reaches 454.53~ms. Such computational dead time forces the robot to execute outdated commands while the environment continues to evolve, leading to collisions in three out of the four real-world scenarios. In comparison, PIER-Flow shows a near-vertical latency CDF in Fig.~\ref{fig:hw_latency_cdf}, indicating highly predictable and low-jitter onboard computation.

\textbf{Conservativeness and Freezing Behavior.}
CBF achieves the lowest latency because it solves only a lightweight reactive QP. However, this low computational cost comes at the expense of overly conservative behavior. When an obstacle blocks the direct path to the goal, CBF tends to decelerate aggressively to maintain safety invariance, which may cause the robot to remain nearly stationary. For example, in the Single Static scenario, CBF requires 12.27~s to complete the task, nearly twice the time of PIER-Flow, which finishes in 6.57~s.

\begin{figure*}[htbp]
	\centering
	\includegraphics[width=1.98\columnwidth]{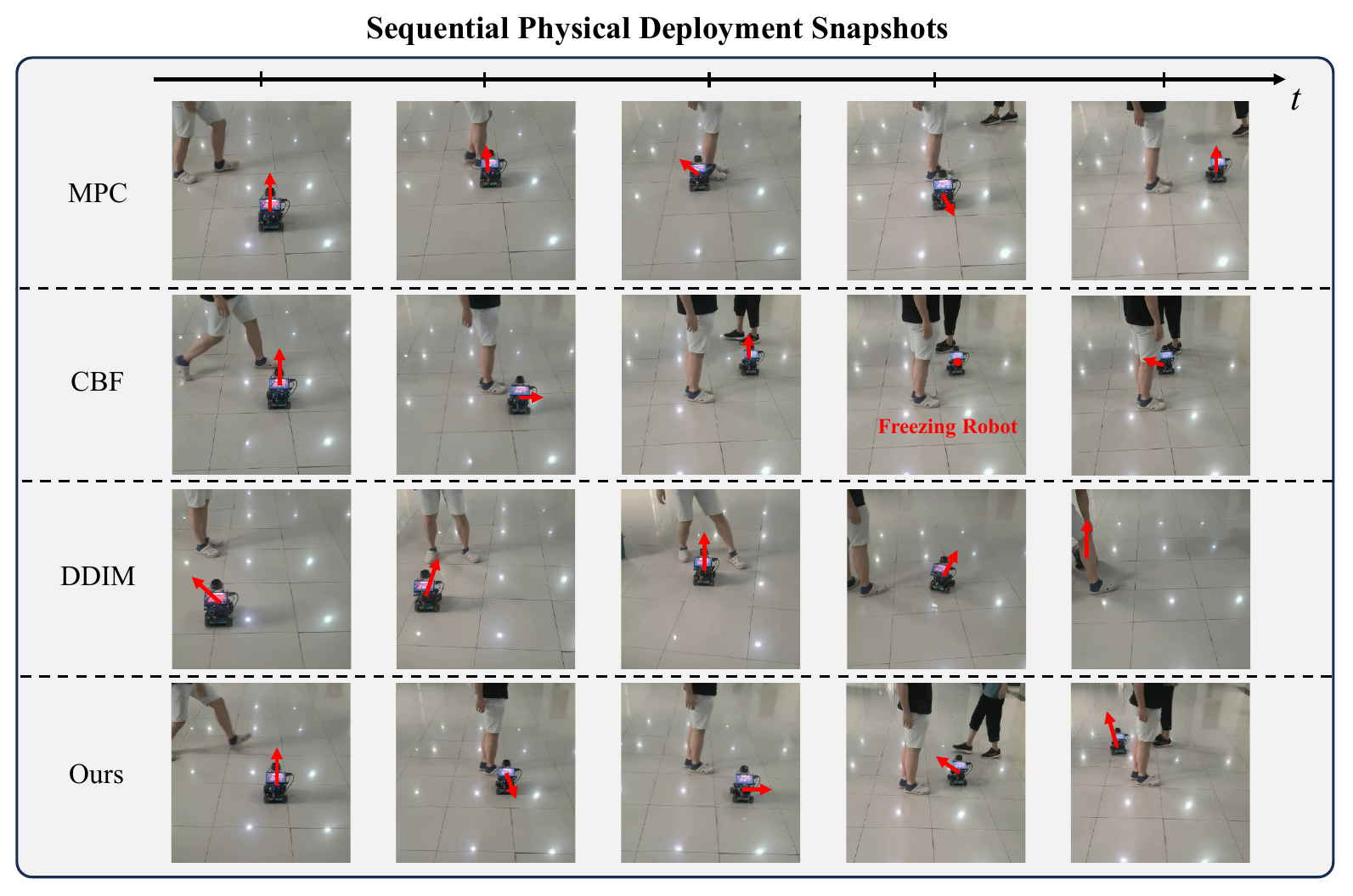}
	\caption{Real-world deployment snapshots in the Multi Dynamic scenario (corresponding to the trajectories in Fig.~\ref{fig:hw_trajectory}). Red arrows indicate the robot's instantaneous velocity direction.}
	\label{fig:hw_snapshots}
\end{figure*}

\textbf{Trajectory Visualization and Analysis.} The recorded trajectories in Fig.~\ref{fig:hw_trajectory} and the deployment snapshots in Fig.~\ref{fig:hw_snapshots} further illustrate the behavioral differences. When confronted with suddenly appearing dynamic obstacles, PIER-Flow generates smooth evasive maneuvers and maintains a non-zero velocity profile until reaching the goal. This behavior results from finite-horizon action-chunk generation, multimodal candidate sampling, and lightweight feasibility selection. By contrast, MPC may fail when solver latency becomes excessive, CBF may stop or slow down under frontal blockage, and DDIM cannot produce timely commands due to its multi-step denoising process.

% [PLACEHOLDER: Insert Figure for "Real-world snapshots from Video"]

\section{Conclusion}
In this paper, we presented PIER-Flow, a lightweight generative navigation framework that combines multimodal action generation, kinematic consistency, and low-latency execution for mobile robots. Trained on state-action sequences generated by an MPC expert, PIER-Flow formulates action-chunk generation as a rectified-flow problem and produces parallel candidates through a single ODE integration step. A differentiable kinematic rollout is incorporated into the training objective to promote physically consistent actions. Together with lightweight feasibility selection and asynchronous action chunking, the proposed framework maintains continuous high-frequency chassis control despite the mismatch between perception, inference, and control rates. 

Extensive evaluations in both simulation and physical edge deployments demonstrated that PIER-Flow consistently outperforms optimization-based methods and traditional diffusion baselines. By maintaining a highly deterministic latency of $\sim$5.3 ms on an onboard GPU, PIER-Flow eliminates the computational latency spikes, as well as the short-sighted ``freezing" behaviors. Future work will explore extending the PIER-Flow framework to incorporate vision-based end-to-end perception and scaling the policy for multi-agent decentralized navigation in unstructured environments.

%\section*{Acknowledgments}

%{\appendix[Proof of the Zonklar Equations]
%Use $\backslash${\tt{appendix}} if you have a single appendix:
%Do not use $\backslash${\tt{section}} anymore after $\backslash${\tt{appendix}}, only $\backslash${\tt{section*}}.
%If you have multiple appendixes use $\backslash${\tt{appendices}} then use $\backslash${\tt{section}} to start each appendix.
%You must declare a $\backslash${\tt{section}} before using any $\backslash${\tt{subsection}} or using $\backslash${\tt{label}} ($\backslash${\tt{appendices}} by itself
% starts a section numbered zero.)}

%{\appendices
%\section*{Proof of the First Zonklar Equation}
%Appendix one text goes here.
% You can choose not to have a title for an appendix if you want by leaving the argument blank
%\section*{Proof of the Second Zonklar Equation}
%Appendix two text goes here.}

 % argument is your BibTeX string definitions and bibliography database(s)
%\bibliography{IEEEabrv,../bib/paper}
%

\bibliographystyle{IEEEtran}
\bibliography{referencerfnav}

%\newpage
%
%\section{Biography Section}
%If you have an EPS/PDF photo (graphicx package needed), extra braces are
% needed around the contents of the optional argument to biography to prevent
% the LaTeX parser from getting confused when it sees the complicated
% $\backslash${\tt{includegraphics}} command within an optional argument. (You can create
% your own custom macro containing the $\backslash${\tt{includegraphics}} command to make things
% simpler here.)
% 
%\vspace{11pt}
%
%\bf{If you include a photo:}\vspace{-33pt}
%\begin{IEEEbiography}[{\includegraphics[width=1in,height=1.25in,clip,keepaspectratio]{fig1}}]{Michael Shell}
%Use $\backslash${\tt{begin\{IEEEbiography\}}} and then for the 1st argument use $\backslash${\tt{includegraphics}} to declare and link the author photo.
%Use the author name as the 3rd argument followed by the biography text.
%\end{IEEEbiography}
%
%\vspace{11pt}
%
%\bf{If you will not include a photo:}\vspace{-33pt}
%\begin{IEEEbiographynophoto}{John Doe}
%Use $\backslash${\tt{begin\{IEEEbiographynophoto\}}} and the author name as the argument followed by the biography text.
%\end{IEEEbiographynophoto}
%
%
%
%
%\vfill

\end{document}